%% file: acl2021.tex
\documentclass[11pt,a4paper]{article}
\usepackage[hyperref]{acl2021}
\usepackage{times}
\usepackage{latexsym}
\usepackage{bclogo}

\usepackage{microtype}

\aclfinalcopy 

\usepackage{bm}
\usepackage{helvet}
\usepackage{courier}
\usepackage{times}
\usepackage{latexsym}
\usepackage{url}
\usepackage{graphicx}
\usepackage{tabularx}
\usepackage{booktabs}
\usepackage{amsmath}
\usepackage{amsfonts}
\usepackage{afterpage}
\usepackage{subfigure}
\usepackage{float}
\usepackage{verbatim}
\usepackage{xcolor}
\usepackage{amssymb}
\usepackage{pifont}
\usepackage{colortbl}
\usepackage{algorithm2e}

\usepackage{microtype}
\usepackage{multirow}
\usepackage{caption}
\usepackage{xcolor}
\definecolor{blue}{RGB}{98, 75, 153}
\definecolor{green}{RGB}{30, 92, 43}
\definecolor{lightgreen}{RGB}{224, 242, 213}
\definecolor{lightred}{RGB}{242, 191, 191}
\definecolor{lightpurple}{RGB}{219, 209, 230}

\newcommand\fb{$^\heartsuit$}
\newcommand\ai{$^\diamondsuit$}
\newcommand\uw{$^\spadesuit$}
\newcommand\msr{$^\clubsuit$}

\title{
\textsc{Go Figure}: A Meta Evaluation of Factuality in Summarization
}

\author{
Saadia Gabriel\uw\thanks{Work done while first author was interning at MSR.} \hspace{10pt}
Asli Celikyilmaz\fb  \hspace{10pt} 
Rahul Jha\msr \hspace{10pt} 
Yejin Choi\uw\ai  \hspace{10pt} 
Jianfeng Gao \msr\\
  \uw Paul G. Allen School of Computer Science \& Engineering,
  University of Washington \\ 
  \fb Facebook AI Research \\
  \msr Microsoft Research  \\
  \ai Allen Institute for Artificial Intelligence \\
  {\tt \{skgabrie,yejin\}@cs.washington.edu} , \tt \{aslic\}@fb.com \\ \tt \{rajh,jfgao\}@microsoft.com}

\date{}

\begin{document}
\maketitle
\begin{abstract}

While neural language models can generate text with remarkable fluency and coherence, controlling for factual correctness in generation remains an open research question. This major discrepancy between the surface-level fluency and the content-level correctness of neural generation has motivated a new line of research that seeks automatic metrics for evaluating the factuality of machine text. 
In this paper, we introduce \textsc{Go Figure},  a \emph{meta-}evaluation framework for evaluating factuality evaluation metrics.  We propose five necessary 
conditions to evaluate factuality metrics 
on diagnostic factuality data across three different summarization tasks.
Our benchmark analysis on ten factuality metrics reveals that our \textit{meta}-evaluation framework provides a robust and efficient evaluation that is extensible to multiple types of factual consistency and standard generation metrics, including QA metrics. It also reveals that while QA metrics generally improve over standard metrics that measure factuality across domains, performance is highly dependent on the way in which questions are generated. 

\end{abstract}

\section{Introduction}

The goal of text generation systems is to produce text that is fluent, coherent, relevant, as well as factually correct. Recent progress in neural approaches to building semantically constrained text generation systems has shown tremendous improvements in this direction \cite{lapata2019,guo-etal-2018-soft,durmus-etal-2020-feqa,wang-etal-2020-asking}. However, an important issue in text generation systems is that they can yield factually inconsistent text, caused by somewhat distorted or fabricated facts about the source text. Especially in document summarization tasks, models that abstract away salient aspects, have been shown to generate text with up to 30\% factual inconsistencies \citep{Kryscinski2019EvaluatingTF,Falke2019RankingGS,zhu2020boosting}.

Commonly used metrics for measuring quality of generated text fail to capture structural aspects of language like negation and poorly correlate with human judgements \cite{Hashimoto2019UnifyingHA,Clark2019SentenceMS,sellam-etal-2020-bleurt}, 
leading to a rapidly progressing search for factuality-driven summarization metrics. 


In this work, we propose GO FIGURE\footnote{\textbf{G}eneral \textbf{O}utline for \textbf{F}actuality \textbf{I}n \textbf{G}enerative \textbf{U}nde\textbf{R}standing \textbf{E}valuation.}, a meta-evaluation framework for assessing the effectiveness of factuality metrics across multiple domains - \textit{extreme summarization}, \textit{multi-sentence news summarization} and the understudied \textit{dialogue summarization} domain. 
Our contributions are as follows: (\textit{i}) a set of diagnostics for measuring sensitivity of metrics to factual inconsistency, (\textit{ii}) a diagnostic evaluation dataset of context/summary pairs for measuring effectiveness of new factuality metrics in a controlled setting, and  (\textit{iii}) an evaluation dataset of summaries generated by transformer-based models \cite{Raffel2019ExploringTL} annotated with types of factual errors. 


\section{Factuality Metric Meta Evaluation}

Since reference summaries may be an incomplete representation of the salient facts 
in a source document or unavailable, we consider factuality in terms of how well candidate summaries are factually grounded with respect to the source document. 


\begin{table*}[!ht]
 \centering
 \resizebox{1\textwidth}{!}{%
 \small
 \begin{tabularx}{\linewidth}{l*{3}{X}}
 \toprule
 Condition &  Definition & \bcloupe Motivation \\ \midrule 
 Boundedness (I) & There exists $ S_r,S_f$ such that \(M(D,S_r) \leq M(D,S_i) \leq M(S_f)\). & In general, the exact factuality level of $S_i$ may be unclear. Metric bounds provide points of comparison. \\ 
Sensitivity (II) & The metric value for $S_i$ should correlate with the level of factuality captured by $S_i$.  & A bounded but insensitive factuality metric may assign higher values to mostly nonfactual or unrelated summaries over summaries that are close to the reference.  \\

Robustness (III) & The metric should be \textit{robust} across types of factual errors. & A metric that is sensitive only to a subset of errors might ignore a significant number of model-generated errors (Figure \ref{fig:real_dist}).  \\

Generality (IV) & The metric should satisfy conditions I,II,III and V across domains.  &  Prior work such as \citet{reiter-belz-2009-investigation} highlight the risk of claiming validity without testing generality.\\
Human Correlation (V)  & 
The metric should \textit{correlate} with human judgements of factuality. & The scoring function $H(D,S_i)$ represented by human evaluation is a gold standard for assessment of generation quality \cite{chaganty-etal-2018-price}, so $M(D,S_i)$ should be an approximation.\\
 \end{tabularx}
 }
 \caption{Details of factuality metric conditions. Here $M$ is a metric scoring function, $D$ is a source document and $S_i$ is a summary.}
 \label{table:conditions}
\end{table*}

We define a set of five conditions for a factual consistency metric $M(D,S_i)$ to measure factuality of a summary $S_i$ with respect to a source document $D$. These conditions are given in Table \ref{table:conditions}.

 \subsection{Testing Factuality Metric Validity} 
 \label{sec:sensitivity}
 
 For the purposes of testing boundedness (Condition I), we define the \textbf{Lower Bound} for a metric $M$ as $M(D, S_r)$ where $D$ is the source document and $S_r$ is a randomly sampled summary from the corpus.\footnote{While this may not be the strictest lower bound in theoretical terms, we consider it appropriate as an empirical lower bound since the content is irrelevant to the document. A single random summary is used.} We define the \textbf{Upper Bound} for the metric as $M(D, S_f)$, where $S_f$ is the reference ground-truth summary. Since our controlled experiments use transformed versions of the reference summary with injected errors, the original reference is guaranteed to be at least as factually consistent as a transformed summary. 
 
\begin{figure}
    \centering
    \includegraphics[width=.9\linewidth]{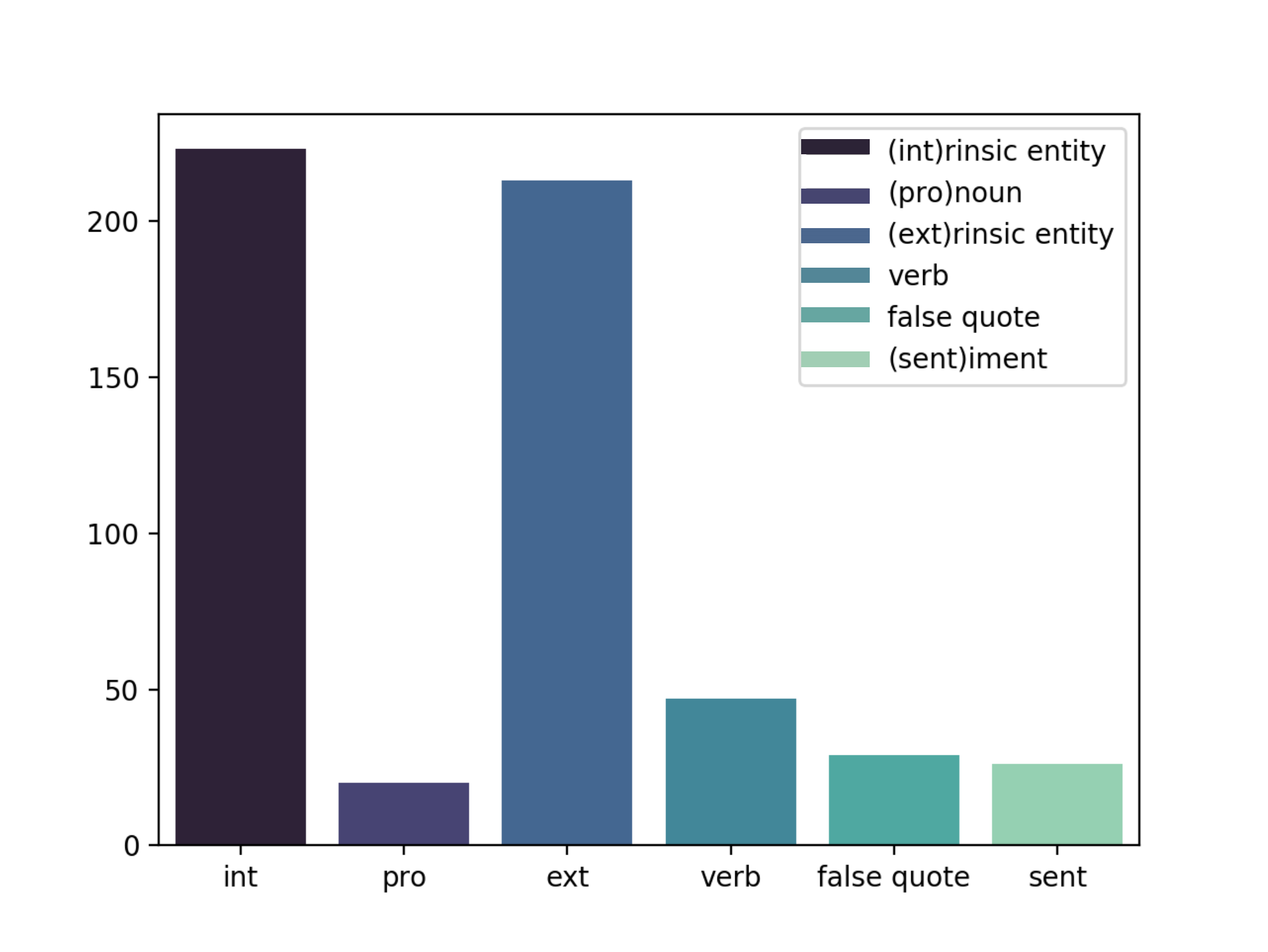}
    \caption{\small Distribution of common factual error types in sampled generated summaries (96.37\% of all errors). We draw from the same error types for our controlled analysis to ensure we match the true distribution of errors. Here \textit{extrinsic entity} refers to entities that did not previously appear in the source, while an \textit{intrinsic entity} appeared in the source.}
    \label{fig:real_dist}
\end{figure}

 To test sensitivity (Condition II), we measure the \textbf{correlation} (Pearson's r) between the factual inconsistency level\footnote{For our experiments, we inject up to a maximum of $x$ errors with $x\in\{1,2,3\}$.} of the summaries (i.e. the number of injected  errors) and the average metric score. Then we measure statistical significance using the \textbf{\textit{p}-value} from a two-tailed hypothesis test. We check whether metrics satisfy robustness and generality (Conditions III and IV) by separately running this analysis over multiple domains and the factual error types shown in Figure \ref{fig:real_dist}. We measure how well metric values match human assessment of factuality by checking the correlation between factual consistency levels determined using manual annotation.
 
 \subsection{Theoretical Cases} 
 
 For Condition I, we scope boundedness to only consider cases that are likely to arise in realistic summarization settings. However, there are hypothetical cases that may have ramifications for metric validity. For example, we expect that $M(D,D) \approx 1 $ and $M(D,\varnothing) \approx 0 $ for a metric $M$ with values in the range $[0,1]$, a document $D$, and an empty string summary $\varnothing$. For non-deterministic metrics, restrictions on variability between runs may also be desired. 
 
\section{Evaluation Datasets}
We evaluate metrics on three datasets: 1-sentence BBC news summaries from the XSUM extreme summarization dataset \cite{xsum-emnlp}, multi-sentence summaries from the CNN/DailyMail dataset \cite{Nallapati2016AbstractiveTS}, and the recently released SAMSUM corpus \cite{Gliwa2019SAMSumCA} consisting of English language conversations written by linguists and aligned multi-sentence summaries.


 \begin{table*}[!ht]
\centering
\scalebox{0.54}{%
\begin{tabular}{l|rr|rrr|rrrrr} \toprule
& \multicolumn{2}{c|}{\underline{\textbf{CLOZE}}} & \multicolumn{3}{c|}{\underline{\textbf{QA}}}  & \multicolumn{5}{c|}{\underline{\textbf{STANDARD and CONTEXTUAL}}} \\
& \textbf{BLANC-Help} & \multicolumn{1}{r|}{\textbf{BLANC-Tune}} &
\textbf{SummaQA-C} & 
\textbf{SummaQA-F1} &
\multicolumn{1}{c|}{\textbf{FEQA}}  & \textbf{R-1} & \textbf{R-2} & \textbf{R-3} & \textbf{R-L} & \textbf{BERTScore}
\\
\midrule
Upper Bound  & 5.99 & 1.73 & 9.64 & 4.48 & 27.87   & 10.61 & 2.56 & 0.72 & 9.32 &   83.76 \\
Level 1  & 5.73 / 5.98 & 1.74 / 1.71 & 9.44 / 9.44 & 3.80 / 4.31 & 23.20  / 26.94  & 10.49 / 10.76 & 2.54 / 2.56 & 0.70 & 9.22 / 9.42  & 83.53 / 83.56\\
Level 2  & 5.46 / 5.99 & 1.59 / 1.78 & 9.27 / 9.35 & 3.40 / 4.22 & 20.05 / 26.55 & 10.40 / 10.86 & 2.51 / 2.54 & 0.69 / 0.68 & 9.16 / 9.49 & 83.36 / 83.38 \\
Level 3  & 5.30 / 5.97 & 1.58 / 1.76 & 9.16 / 9.23 & 3.13 / 4.14 & 15.81 / 26.06 & 10.33 / 10.92 & 2.49 / 2.52 & 0.69 / 0.67  & 9.10 / 9.55 & 83.21 / 83.26 \\ 
Lower Bound  & 0.51 & -0.14 & 1.28 & 0.26 & 1.18  & 5.44 & 0.39 & 0.01 & 4.94 & 80.08   \\
\midrule
Correlation  & -0.99 / -0.61 & -0.88 / \textbf{\colorbox{lightpurple}{0.69}} & -0.99 / \textbf{\colorbox{lightgreen}{-1.00}} & -0.99 / \textbf{\colorbox{lightgreen}{-1.00}} & \textbf{\colorbox{lightgreen}{-1.00}}   & \textbf{\colorbox{lightgreen}{-1.00}} / \textbf{\colorbox{lightpurple}{0.98}} & -0.97 / \textbf{\colorbox{lightgreen}{-1.00}} & \textbf{\colorbox{lightred}{-0.87}} / \textbf{\colorbox{lightgreen}{-1.00}} & \textbf{\colorbox{lightgreen}{-1.00}} / \textbf{\colorbox{lightpurple}{1.00}} & \textbf{\colorbox{lightgreen}{-1.00}} \\ 
p-value  & 0.09 / 0.59 & 0.32 / 0.51 & 0.07 / 0.05* & 0.07 / 0.03* & 0.05* / 0.04*  & 0.03* / 0.10 & 0.16 / 0.05* & 0.33 / 0.05* & $<$0.01** / 0.02* & 0.02* / 0.06\\
\bottomrule
\end{tabular}}
\vspace{-1ex}
\caption{\small Results of simulated factual error data experiments (\textbf{XSUM}, average of 5 runs, **=significant for p $\leq$ .01, *=significant for p $\leq$ .05). \textbf{For cells with ($\cdot$/$\cdot$), results for entity errors are reported on the left, results for non-entity errors are reported on the right.} The details for the upper/lower bounds, \textit{p}-value and correlation measures are explained in $\S$\ref{sec:sensitivity}. For sensitivity to factual consistency and correlation w/ factuality levels, we highlight the best-performing and lowest-performing metrics in \textbf{\textcolor{green}{green}} and \textbf{\textcolor{red}{red}} respectively. For cases where metric values are invalid (e.g. the metric values increase as factuality decreases), we highlight in \textbf{\textcolor{violet}{purple}}. }
\label{table:xsum}
\end{table*}

\begin{table*}[!ht]
\centering
\scalebox{0.54}{%
\begin{tabular}{l|rr|rrr|rrrrr} \toprule
& \multicolumn{2}{c|}{\underline{\textbf{CLOZE}}} & \multicolumn{3}{c|}{\underline{\textbf{QA}}} & \multicolumn{5}{c|}{\underline{\textbf{STANDARD and CONTEXTUAL}}} \\
& \textbf{BLANC-Help} & \multicolumn{1}{r|}{\textbf{BLANC-Tune}} &
\textbf{SummaQA-C} & 
\textbf{SummaQA-F1} &
\multicolumn{1}{c|}{\textbf{FEQA}}   & \textbf{R-1} & \textbf{R-2} & \textbf{R-3} & \textbf{R-L} & \textbf{BERTScore}
\\
\midrule
Upper Bound  & 7.60 & 5.79 & 13.82 & 10.87 & 37.56  & 14.33 & 8.08 & 4.75 & 13.83 & 84.36   \\
Level 1  & 7.29 / 7.50  & 5.56 / 5.69 & 13.30 / 13.53  & 9.58 / 10.63  & 33.35  / 36.64 & 14.11 / 14.37 &  7.78 / 7.91 & 4.51 / 4.57 & 13.60 / 13.84 & 84.13 / 84.20\\
Level 2  & 7.03 / 7.43  & 5.43 / 5.58  & 12.93 / 13.24 & 8.53 / 10.38 & 28.46 / 36.13 &  13.95 / 14.38 & 7.55 / 7.75 & 4.32 / 4.40 & 13.44 / 13.85 & 83.94 / 84.04 \\
Level 3  & 6.72 / 7.38 & 5.23 / 5.53  & 12.54 / 13.04 & 7.54 / 10.26  & 25.12 / 35.63  & 13.82 / 14.38 & 7.35 / 7.62 & 4.14 / 4.27 & 13.29 / 13.85 & 83.77 / 83.90\\ 
Lower Bound  & -0.67 & -0.19 & 1.61 & 0.12 & 0.58 & 5.85 & 0.47 & 0.02 & 5.55 & 78.16  \\
\midrule
Correlation  & \textbf{\colorbox{lightgreen}{-1.00}} / -0.99 &  \textbf{\colorbox{lightred}{-0.99}} / -0.97 & \textbf{\colorbox{lightgreen}{-1.00}} / \textbf{\colorbox{lightgreen}{-1.00}} & \textbf{\colorbox{lightgreen}{-1.00}} / -0.98 & \textbf{\colorbox{lightred}{-0.99}} / \textbf{\colorbox{lightgreen}{-1.00}} &  \textbf{\colorbox{lightgreen}{-1.00}} /  \textbf{\colorbox{lightpurple}{0.96}} & \textbf{\colorbox{lightgreen}{-1.00}} & \textbf{\colorbox{lightgreen}{-1.00}} & \textbf{\colorbox{lightgreen}{-1.00}} / \textbf{\colorbox{lightpurple}{0.91}} & \textbf{\colorbox{lightgreen}{-1.00}}\\ 
p-value  & 0.03* / 0.08 & 0.07 / 0.17 & 0.01** / 0.06  & 0.01** / 0.13 & 0.07 / $<$0.01** & 0.04* / 0.17 & 0.02* / 0.04*  & $<$0.01** / 0.04* & 0.03* / 0.27 & 0.01** / 0.02*\\
\bottomrule
\end{tabular}}
\vspace{-1ex}
\caption{\small Results of simulated factual error data experiments (\textbf{CNNDM}, average of 5 runs). (See Table~\ref{table:xsum} caption for details.)}
\label{table:cnndm}
\end{table*}

\begin{table*}[!ht]
\centering
\scalebox{0.54}{%
\begin{tabular}{l|rr|rrr|rrrrr} \toprule
& \multicolumn{2}{c|}{\underline{\textbf{CLOZE}}} & \multicolumn{3}{c|}{\underline{\textbf{QA}}}  & \multicolumn{5}{c|}{\underline{\textbf{STANDARD and CONTEXTUAL}}} \\
& \textbf{BLANC-Help} & \multicolumn{1}{r|}{\textbf{BLANC-Tune}} &
\textbf{SummaQA-C} & 
\textbf{SummaQA-F1} &
\multicolumn{1}{c|}{\textbf{FEQA}}   & \textbf{R-1} & \textbf{R-2} & \textbf{R-3} & \textbf{R-L} & \textbf{BERTScore}
\\
\midrule
Upper Bound  & 15.23 & 10.13 & 13.83 & 17.23 & 55.36   & 26.55 & 8.24 & 4.07 & 25.06  & 84.60  \\
Level 1  & 13.97 / 15.03   & 9.00 / 9.47  & 13.48 / 13.52   & 15.00 / 16.71 & 45.31 / 54.25  & 25.31 / 26.18 & 7.85 / 7.86 & 3.84 / 3.73 & 23.91 / 24.69  & 84.42 / 84.38\\
Level 2  & 12.87 / 15.01  & 8.36 / 9.46  & 13.16 / 13.26 & 12.26 / 16.50 & 37.01 / 53.10 & 24.27 / 25.86  & 7.60 / 7.59 & 3.68 / 3.50  & 22.99 / 24.38  & 84.28 / 84.19 \\
Level 3  & 12.02 / 14.93 & 7.74 / 9.36   & 12.99 / 13.21 & 10.12 / 16.24  & 29.62 / 52.34 & 23.23 / 25.58 & 7.32 / 7.36  & 3.48 / 3.35  & 22.01 / 24.12 & 84.13 / 84.07\\ 
Lower Bound  & 0.92 & -0.53 & 7.86 & 0.10 & 0.55   & 5.33 & 0.23  & 0.01 & 5.09 & 80.79  \\
\midrule
Correlation  & \textbf{\colorbox{lightgreen}{-1.00}} / -0.96  &  \textbf{\colorbox{lightgreen}{-1.00}} / \textbf{\colorbox{lightred}{-0.91}} & \textbf{\colorbox{lightred}{-0.99}} / -0.94 & \textbf{\colorbox{lightgreen}{-1.00}}  & \textbf{\colorbox{lightgreen}{-1.00}} / -0.99  & \textbf{\colorbox{lightgreen}{-1.00}} &   \textbf{\colorbox{lightgreen}{-1.00}} & \textbf{\colorbox{lightgreen}{-1.00}} / -0.99  & \textbf{\colorbox{lightgreen}{-1.00}} & \textbf{\colorbox{lightgreen}{-1.00}}/-0.99 \\ 
p-value  & 0.05* / 0.18  & 0.01** / 0.28 & 0.11 / 0.23  & 0.05* & 0.02* / 0.07  & $<$0.01** / 0.03* & 0.03*  & 0.05* / 0.08  & 0.01** / 0.04* & 0.01** / 0.07 \\
\bottomrule
\end{tabular}}
\vspace{-1ex}
\caption{\small Results of simulated factual error data experiments (\textbf{SAMSUM}, average of 5 runs). (See Table~\ref{table:xsum} caption for details.)}
\label{table:samsum}
\end{table*}

\subsection{Diagnostic Datasets} 

To test the ability of proposed metrics to fulfill our predefined conditions, we set up two diagnostic datasets consisting of (\textit{i}) transformed reference summaries with simulated factuality errors that allow us to induce and measure factuality levels in a controlled setting and (\textit{ii}) summaries generated by state-of-the-art transformer summarization models that allows us to measure the effectiveness of metrics in a real data setting. We sample 500 source / summary pairs for each domain.\footnote{See the Appendix for details of linguistic feature extraction for injecting errors.} 


\subsubsection{Model-Generated Datasets}
 \label{sec:generated}
 
 In order to observe how metrics perform on machine-generated summaries, we generate summaries from fine-tuned T5 encoder-decoder summarization models \cite{Raffel2019ExploringTL} that was pretrained on news summarization data. We generate summary text using either beam search or sample-based decoding strategies. We then annotate the generated summaries for fine-grained factual errors using the types in Figure \ref{fig:real_dist} to create a hand-curated factual consistency diagnostic dataset. 
 

\section{Factuality Metrics for Evaluation}

We mainly focus on meta-evaluating most recently proposed factual consistency metrics which use two types of proxy natural language understanding (NLU) objectives aimed at implicitly capturing factuality in generated text: \textit{question-answering} (QA) and a masked token prediction \textit{cloze task}. For QA we evaluate using SummaQA \cite[which uses QA pairs from the source,][]{Scialom2019AnswersUU} and FEQA \cite[which uses QA pairs from the summary,][]{durmus-etal-2020-feqa}, while for the cloze task setting we use BLANC-Help and BLANC-Tune \cite[see the appendix for details of metrics]{Vasilyev2020FillIT}. We also measure the \textit{factual-awareness} of BERTScore \cite{Zhang2020BERTScoreET}, a summarization metric that is aimed primarily at improving coherency rather than factual consistency, and standard summarization evaluation metrics (e.g. ROUGE \cite{lin-2004-rouge}).

\section{Meta-Analysis of Factuality Metrics}

\subsection{Controlled Data Experiments} 
\label{sec:gen_exp}

We provide the results of the sensitivity analysis over our controlled data on the XSUM domain in Table~\ref{table:xsum}, on CNNDM in Table~\ref{table:cnndm} and on SAMSUM in Table~\ref{table:samsum}. 
Our analysis reveals that QA metrics, ROUGE-(2/3) and BERTScore generally perform well at evaluating factuality. In contrast, ROUGE-(1/L) are frequently invalid as factuality metrics (Tables \ref{table:xsum} and \ref{table:cnndm}), and the performance of Cloze metrics varies across domains (BLANC-Tune is invalid on XSUM, but does fairly well on other domains). Also, performance of metrics tends to be much lower on news domains when we consider non-entity-based errors with the exception of QA-based metrics, ROUGE-(2/3) and BERTScore, indicating that while factuality and standard metrics are fairly attuned to changes in factual consistency that relate to entity-based errors, they are less robust to other types of factual errors. 

\subsection{Comparison with Human Evaluation of Model Generations}


 
We find that metrics displaying invalid behavior on controlled data (for instance assigning higher metric values to more factually inconsistent summaries on XSUM in Table \ref{table:xsum}) also display this invalid behavior in model generations (Table \ref{table:corr1}). This indicates that meta-evaluation with controlled data is effective as a diagnostic tool for finding weak factuality metrics, and follows our intuition that non-entity errors, while frequently produced by abstractive summarization models, are difficult for standard summarization metrics to identify. When considering better-performing factuality metrics identified by the controlled error analysis, we find that the controlled data analysis is generally able to identify better-performing metrics (SummaQA, ROUGE-(2/3) and BERTScore) for XSUM with the exception of FEQA (FEQA metric performs well on XSUM controlled analysis (Table \ref{table:xsum}), but only approaches this performance on SAMSUM when we consider human eval). The strong overall performance of ROUGE-3 is consistent with the findings of \cite{Fabbri2020SummEvalRS} on CNNDM, our work confirms that this metric is more consistently correlated with factuality than other ROUGE variations across domains.

\begin{table}[t]
 \centering
 \resizebox{1\linewidth}{!}{%
 \begin{tabular}{ c |  c l | c l } 
 \toprule
 \textbf{Metric} & \textbf{XSUM} & & \textbf{SAMSUM} \\ \cmidrule{2-5}
 & \textbf{Corr} (- $\leftarrow$) & \textbf{\textit{p}-value} &  \textbf{Corr} (- $\leftarrow$) & \textbf{\textit{p}-value}\\
 \midrule
 BLANC-Help & 0.04  & 0.55 & -0.01 & 0.82 \\
 BLANC-Tune & 0.00 & 0.98 & -0.03 & 0.64\\
 SummaQA-C &  -0.11 & 0.11 & -0.09 & 0.18 \\
 SummaQA-F1 & -0.12 & 0.07 & -0.14 & 0.03* \\
 FEQA & 0.04 & 0.57 & -0.03 & 0.69 \\
 R-1 & 0.07 & 0.19 & 0.01 & 0.82  \\
 R-2 & -0.10 & 0.15 & -0.03 & 0.59\\
 R-3 & -0.12 & 0.07 & -0.09 & 0.18\\
 R-L & 0.07 & 0.13 & 0.01 & 0.83\\
 BERTScore & -0.17 & 0.01** & 0.03 & 0.64  \\
 \bottomrule
 \end{tabular}
 }
 \caption{\small Correlation (Corr) for 250 annotated XSUM and 250 SAMSUM generated summaries with fine-grained labeling. The arrow next to ``Corr" indicates the direction of a correct correlation.}
 \label{table:corr1}
 \end{table}
 

\section{Related Work} 

Prior work concerning evaluation of automatic metrics and human evaluation for NLG systems has mainly focused on general analysis of output quality or coherence and fluency \cite{callison-burch-etal-2007-meta,graham-2015-evaluating,Fabbri2020SummEvalRS}, rather than factuality. Recent efforts by NLP researchers have drawn attention to the issue of factual errors and hallucinations in the output of neural summarization models \cite{Cao2018FaithfulTT,Massarelli2019HowDS,Zhao2020ReducingQH,falke-etal-2019-ranking,Goodrich2019AssessingTF,elikyilmaz2020EvaluationOT}. A number of works have highlighted the effectiveness of QA and cloze task objectives for evaluating or improving factuality on specific domains \cite{eyal-etal-2019-question,huang-etal-2020-knowledge}. We aim to evaluate these metrics more broadly, and consider a wider range of domains (notably dialogue).

\subsection{Discussion of Meta Evaluation and Conclusion}

Our analyses show that in contrast to prior work on factual consistency that mostly concentrated on one specific domain and dataset, our GO FIGURE framework is effective at evaluating sensitivity and validity of factual consistency metrics with only reference summaries, rather than requiring computationally intensive testing across summarization model variants to identify metric strengths and shortcomings.

We highlight the following key points from experiments run using \textit{meta}-evaluation:


\textbf{Standard summarization metrics are not always valid measures of factuality.} ROUGE-1 and ROUGE-L fail to accurately measure factual inconsistency across domains in our controlled analysis. The ROUGE-L results raise the question of context \textit{relevance}. While ROUGE-L takes into account more context than other ROUGE variations, this context may not be relevant for assessing factuality. For example, swapping ``decreased" for ``increased" dramatically changes the meaning in the summary \textit{``Scotland's renewable energy output increased by 45\% in the first quarter of this year, compared with the same period last year."}, but ROUGE-L is not affected. Despite the frequent use of ROUGE-L as a more contextual measure, prior work has also noted that ROUGE-N outperforms ROUGE-L \cite{rankel-etal-2013-decade, Fabbri2020SummEvalRS}. 

\textbf{Analysis on human annotated data is still necessary as an upper-bound on \textit{meta}-evaluation quality.} While BLANC-Help, FEQA metric and BERTScore values decrease with factual inconsistency on controlled data, the metrics may sometimes be positively correlated with factual inconsistency on generated data. This emphasizes the importance of a expert curated test set as part of the GO FIGURE meta evaluation for the most rigorous testing. \textbf{A question-answering objective is promising for measuring factual consistency across domains, but effectiveness depends on the question.} While QA metrics can perform well at measuring factual consistency of generated summaries, our \textit{meta}-evaluation reveals this is dependent on the way in which questions are asked. While both QA metrics use SQuAD-based systems \cite{Rajpurkar2016SQuAD10}, asking questions from the source rather than the summary is most robust across domains. This opens the door to metrics based on more contextual QA like commonsense \cite{Shwartz2020UnsupervisedCQ}.


We will release our \textit{meta}-evaluation framework and diagnostic datasets to aid in development of effective summarization factuality metrics. In future work, summary meta-metric results (e.g. correlation on simulated data) could be used as rewards for reinforcement learning driven approaches to training factuality metrics. 


\section{Ethics and Broader Impact Statement}

Ethical considerations involving our \textit{meta}-evaluation framework primarily revolve around human evaluation. News articles and dialogues may contain references to distressing events or abnormal social behavior. All our expert annotators voluntarily took part in the human evaluation with prior knowledge of the type of content being evaluated. Crowd-sourced human evaluation trials were conducted under an IRB exemption. 

Our work outlines a simple and effective approach for evaluating factuality metrics in summarization. This can aid in development of more robust and sensitive factuality metrics to accurately evaluate the factual correctness of generative models. This is key as improvement in the coherency of models accelerates, potentially leading to generations that appear to be high quality while containing factual inaccuracies. Our framework could also evaluate factuality metrics for use in identifying human-written errors, mitigating potential spread of misinformation. 

\section*{Acknowledgments}

The authors thank Yichen Jiang and Shiyue Zhang for feedback on implementation, Hannah Rashkin and Tom McCoy for help with MSR GPU clusters, Rowan Zellers and Elizabeth Clark for pointers to related work, as well as other members of the UW NLP, MSR AI and MSR MSAI communities for helpful comments. 

\bibliography{anthology,acl2020}
\bibliographystyle{acl_natbib}

\clearpage

\appendix

\section{Appendices}
\label{sec:appendix}

\input{appendix.tex}

\end{document}

%% file: appendix.tex
\subsection{Additional Details of Datasets}

We provide dataset statistics for each of our domains in Table \ref{table:domains}. 

 \begin{table}[t]
 \centering
 \resizebox{1\linewidth}{!}{%
 \begin{tabular}{  l  l l l l } 
 \toprule
 \textbf{Dataset}  & \textbf{Train} & \textbf{Dev} & \textbf{Test} & \textbf{Domain}\\
 \midrule
  XSUM & 204,045 & 11,332 & 11,334 & Short news \\
  CNNDM & 287,227 & 13,368 & 11,490 & Long news \\
  SAMSUM & 14,732 & 818 & 819 & Dialogues \\
 \bottomrule
 \end{tabular}
 }
 \caption{\small Summarization domains for evaluation.}
 \label{table:domains}
 \end{table}

\begin{table*}[t]
 \centering
 \resizebox{\linewidth}{!}{%
 \begin{tabular}{  l  c l l } 
 \toprule
 \textbf{Reference}  & \textbf{Type} & \textbf{Description} &  \textbf{Example}\\
 \midrule
   \textcolor{green}{Irish} Taoiseach (PM) Leo Varadkar has engaged in & & An entity appearing in the & \textcolor{red}{Canadian} Taoiseach (PM) Leo Varadkar has engaged in\\ 
   some ``sock diplomacy" in his first meeting with & Intrinsic entity error & source document is used & some ``sock diplomacy" in his first meeting with \\
   \textcolor{green}{Canadian} Prime Minister Justin Trudeau in Dublin. &   (\textbf{int}) & incorrectly. & \textcolor{red}{Irish} Prime Minister Justin Trudeau in Dublin. \\ \midrule 
   \textcolor{green}{Irish} Taoiseach (PM) Leo Varadkar has engaged in & & An entity appearing in & \textcolor{red}{French} Taoiseach (PM) Leo Varadkar has engaged in\\ 
   some ``sock diplomacy" in his first meeting with & Extrinsic entity error & the candidate summary does & some ``sock diplomacy" in his first meeting with \\
   Canadian Prime Minister Justin Trudeau in Dublin. &  (\textbf{ext}) & not appear in the source document. & Canadian Prime Minister Justin Trudeau in Dublin. \\ \midrule
   Irish Taoiseach (PM) Leo Varadkar has engaged in & & A pronoun in the candidate summary & Irish Taoiseach (PM) Leo Varadkar has engaged in\\ 
   some ``sock diplomacy" in \textcolor{green}{his} first meeting with & Pronoun error & is used incorrectly. & some ``sock diplomacy" in \textcolor{red}{her} first meeting with \\
   Canadian Prime Minister Justin Trudeau in Dublin. & (\textbf{pro}) & For example, (her/she instead of him/he). & Canadian Prime Minister Justin Trudeau in Dublin. \\ \midrule
   Irish Taoiseach (PM) Leo Varadkar \textcolor{green}{has engaged in} & & There are verb negations in & Irish Taoiseach (PM) Leo Varadkar \textcolor{red}{has not engaged in} \\ 
   some ``sock diplomacy" in his first meeting with & Negation error & the candidate summary that & some ``sock diplomacy" in his first meeting with \\
   Canadian Prime Minister Justin Trudeau in Dublin. & (\textbf{verb}) & contradict the source document. & Canadian Prime Minister Justin Trudeau in Dublin. \\ \midrule
   People who have been prescribed \textcolor{green}{powerful} anxiety & & An adjective or adverb appearing & People who have been prescribed \textcolor{red}{weak} anxiety\\ 
    or pain relief drugs are being warned about a new & Sentiment error & in the candidate summary & or pain relief drugs are being warned about a new\\ 
    drug-driving law. &  (\textbf{sent}) & contradicts the source document. & drug-driving law. \\ \midrule
 \bottomrule
 \end{tabular}
 }
 \caption{\small Table of possible factual errors.}
 \label{table:fact_errors}
 \end{table*}

\subsection{Evaluation Metric Details} 

\indent \textbf{QA-Based Quality Score.} Given a source or reference document $D$ and candidate summary $S_i$, QA-based evaluation metrics assign a generation quality score to $S_i$ to measure the ability of a QA system by accurately answering questions generated from $D$ or $S_i$. We use the SummaQA \cite{Scialom2019AnswersUU} and FEQA \cite{durmus-etal-2020-feqa} metrics. For the SummaQA metric, questions are generated from the source document $D$ and the candidate summary $S_i$ is used as input to the QA system. Alternatively, FEQA generates questions from $S_i$ and uses $D$ to answer these questions. 

The generation quality score is typically the aggregated $F_1$ score measuring the similarity between ground-truth answers for questions generated from $D$ and the answers predicted by the QA system. SummaQA also generally includes the aggregated model confidence probabilities for predictions. 

\indent \textbf{Masked LM Prediction (Cloze Task) Score.} Given a source document $D$ and candidate summary $S_i$, Cloze-based evaluation metrics assign a generation quality score to $S_i$ by measuring the ability of a NLU system to accurately predict masked tokens in the source document, given access to the information in $S_i$. We use two variants of BLANC \cite{Vasilyev2020FillIT}, BLANC-Help and BLANC-Tune. BLANC-Help uses both $D$ and $S_i$ as input to a pretrained masked token prediction model, while BLANC-Tune only uses $D$ as input to a model that has been finetuned on the candidate summary. Both metrics are aimed at capturing fluency, informativeness and factual correctness of summaries. 

\indent \textbf{Semantic Similarity.} Semantic similarity metrics measure the overlap between contextual embeddings of a source or reference document $D$ and candidate summary $S_i$. We use BERTScore \cite{Zhang2020BERTScoreET}, which has been shown to correlate better with human judgements of coherency than standard summarization metrics and similarly to \textit{n}-gram metrics on factual consistency of CNNDM summaries \cite{wang-etal-2020-asking}.

\indent \textbf{Lexical Overlap.} Finally, we test ROUGE \cite{lin-2004-rouge}, which is the standard metric used for evaluating summarization. ROUGE measures the \textit{n}-gram overlap between a source or reference document $D$ and candidate summary $S_i$. We evaluate results using ROUGE-1 and ROUGE-2, as well as ROUGE-L, which measures longest common sub-sequence overlap. We follow prior work that considered ROUGE in factual consistency evaluations \cite{wang-etal-2020-asking}, though it has also been previously noted that ROUGE can underweight good summarization examples \cite{Novikova2017WhyWN}.

\subsection{Simulated Data Transformations}

We inject errors into reference summaries by first using a part-of-speech tagging model and named entity recognition system (spaCy)\footnote{\url{https://spacy.io/}} to extract entities, verbs, and adjectives from these summaries. For each named entity, we keep track of the label type (e.g. ORG, GPE, etc). All datasets are comprised of English language articles or dialogues and summaries, and we use the spaCy English NLP models. 

\textbf{Intrinsic entity errors.} To inject intrinsic entity errors into a summary $S$, we construct a dictionary of all unique entities appearing in the source document for $S$ \textbf{only}, organized by entity label type. We then swap a random entity in the reference summary for a different entity of the same label type in the constructed dictionary. 

\textbf{Extrinsic entity errors.} For extrinsic entity errors, we use the same dictionary construction for all unique entities appearing in \textbf{all} the corpus source documents. To change a random adjective, we use WordNet \cite{10.1145/219717.219748} to obtain the synsets for that adjective and swap the adjective for its antonym. 

\textbf{Pronoun entity errors.} Pronoun errors are introduced with a preset list of commonly used pronouns. We randomly extract a pronoun set (e.g. she/her) from the text using the preset list and swap it with another random pronoun set (e.g. he/him).

\textbf{Verb Negation.} We use a rule-based system for verb negation based on verb tense, and predict tense based on the suffix and preceding words. 

We note that injecting a certain level of error into a summary will have varying effects depending on the average length of summaries for a corpus. We use the same methodology for each corpus to maintain consistency, but future work may explore length-controlled error injection based on the objectives of the evaluation.  

\subsection{Metric Implementation Details}

For all metrics, we use the publicly shared implementations. Due to BERT context size constraints, we limit the length of document input sentences to 400 tokens for BLANC variants. We use Roberta-large for BERTScore. 

\subsection{T5 Training}

We fine-tune the T5-base model (220M parameters) trained on news summaries for each domain using the AdaFactor optimizer \cite{Shazeer2018AdafactorAL} with a learning rate of 0.001 and a batch size of 8. The learning rate was tuned using ROUGE score on a dev set, and we experimented with learning rates in the range of [0.01,0.0001]. All other hyperparameters follow from the original T5 paper. Best performing models were trained using one random seed on NVIDIA V100 GPUs.

\subsubsection{Human Annotation Layout} 

For human annotation of factual consistency in summaries, we show the source document, reference summary and a candidate summary that should be assessed for factuality. We then ask a factuality question with three choices:
\begin{itemize}
    \item Yes (i.e. the summary is factual) 
    \item No (i.e. the summary contains factual inconsistencies)
    \item Not Sure (i.e. the summary is too incoherent to judge)
\end{itemize}

\begin{figure*}
    \centering
    \includegraphics[width=1\linewidth]{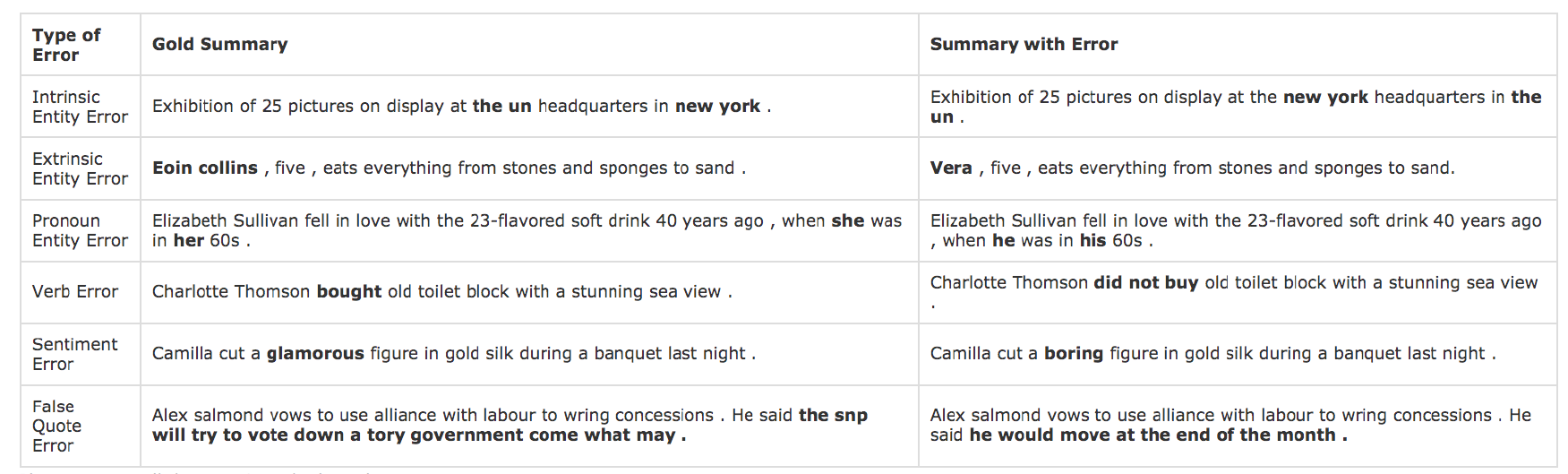}
    \caption{\small Examples of factual errors given in annotation task.}
    \label{fig:mturk_examples}
\end{figure*}

If a summary is judged to be factually incorrect, annotators are allowed to select the number and type of errors they observe using a predefined list of factual errors. A screenshot of the error types and examples shown in the annotation task is given in Figure \ref{fig:mturk_examples}. For less obvious cases of factual inconsistency (for example when summaries contain locations or political figures that require regional background knowledge), we check factuality using external knowledge bases to ensure correctness of annotation. We also adhere to a strict binary notion of factuality in deciding cases where summaries are imprecise but ambiguous in terms of correctness, opting to label these summaries as factually inaccurate. If summaries are completely incoherent, we treat these summaries as having the highest level of factual inconsistency. 

We validated the effectiveness of the setup by computing inter-annotator agreement of in-house expert annotators for 30 XSUM summaries. We achieve ``fair" agreement of Krippendorff's $\alpha$ = 0.32 with 3 annotators and ``moderate" agreement of $\alpha$ = 0.44 with 2 annotators \cite{Landis1977TheMO,Ageeva2015EvaluatingMT}. The remaining annotations are done by one in-house expert annotator. 

 \begin{table*}[t]
 \centering
 \resizebox{.8\linewidth}{!}{%
 \begin{tabular}{  l  c c c c } 
 \toprule
 \textbf{Dataset}  & \textbf{Level 1 Avg.} & \textbf{Level 2 Avg.} & \textbf{Level 3 Avg.} & \textbf{Avg. \% Transformed}\\
 & & & & (L1/L2/L3/All)\\
 \midrule
  XSUM (Entity) & 0.59 & 1.14 & 1.61 & 58.84 / 76.44 / 86.28 / 73.85\\
  XSUM (Non-Entity) & 0.48 & 0.93 & 1.28 & 48.32 / 74.00 / 85.40 / 69.24\\
  CNNDM (Entity) & 0.75 & 1.48 & 2.17 & 74.92 / 85.68 / 94.48 / 85.03 \\
  CNNDM (Non-Entity) & 0.50 & 1.05 & 1.62 & 79.44 / 93.32 / 97.04 / 89.93\\
  SAMSUM (Entity) & 0.59 & 1.16 & 1.70  & 58.96 / 77.32 / 87.56 / 74.61 \\
  SAMSUM (Non-Entity) & 0.49 & 0.91 & 1.28  & 48.52 / 72.80 / 84.12 / 68.48 \\
 \bottomrule
 \end{tabular}
 }
 \caption{\small Analysis of simulated diagnostic dataset (we average across 5 different sets (runs) of randomized transformations for the same 500 reference summaries). We provide results for the average number of induced factuality errors for factual inconsistency level 1 (L1), level 2 (L2) and level 3 (L3), as well as the percentage (\%) of summaries that were transformed for each level and across all levels (All). We split the diagnostic dataset into two subsets based on whether simulated errors are related to entities (Entity) or non-entity changes like verb negation (Non-Entity). }
 \label{table:diagnostic}
 \end{table*}

%% file: acl2021.bbl
\begin{thebibliography}{39}
\expandafter\ifx\csname natexlab\endcsname\relax\def\natexlab#1{#1}\fi

\bibitem[{Ageeva et~al.(2015)Ageeva, Forcada, Tyers, and
  P{\'e}rez-Ortiz}]{Ageeva2015EvaluatingMT}
E.~Ageeva, M.~Forcada, Francis~M. Tyers, and Juan~Antonio P{\'e}rez-Ortiz.
  2015.
\newblock Evaluating machine translation for assimilation via a gap-filling
  task.
\newblock In \emph{EAMT}.

\bibitem[{Callison-Burch et~al.(2007)Callison-Burch, Fordyce, Koehn, Monz, and
  Schroeder}]{callison-burch-etal-2007-meta}
Chris Callison-Burch, Cameron Fordyce, Philipp Koehn, Christof Monz, and Josh
  Schroeder. 2007.
\newblock \href {https://www.aclweb.org/anthology/W07-0718} {(meta-) evaluation
  of machine translation}.
\newblock In \emph{Proceedings of the Second Workshop on Statistical Machine
  Translation}, pages 136--158, Prague, Czech Republic. Association for
  Computational Linguistics.

\bibitem[{Cao et~al.(2018)Cao, Wei, Li, and Li}]{Cao2018FaithfulTT}
Ziqiang Cao, Furu Wei, W.~Li, and Sujian Li. 2018.
\newblock Faithful to the original: Fact aware neural abstractive
  summarization.
\newblock In \emph{AAAI}.

\bibitem[{Celikyilmaz et~al.(2020)Celikyilmaz, Clark, and
  Gao}]{elikyilmaz2020EvaluationOT}
Asli Celikyilmaz, Elizabeth Clark, and Jianfeng Gao. 2020.
\newblock Evaluation of text generation: A survey.
\newblock \emph{ArXiv}, abs/2006.14799.

\bibitem[{Chaganty et~al.(2018)Chaganty, Mussmann, and
  Liang}]{chaganty-etal-2018-price}
Arun Chaganty, Stephen Mussmann, and Percy Liang. 2018.
\newblock \href {https://doi.org/10.18653/v1/P18-1060} {The price of debiasing
  automatic metrics in natural language evalaution}.
\newblock In \emph{Proceedings of the 56th Annual Meeting of the Association
  for Computational Linguistics (Volume 1: Long Papers)}, pages 643--653,
  Melbourne, Australia. Association for Computational Linguistics.

\bibitem[{Clark et~al.(2019)Clark, Celikyilmaz, and
  Smith}]{Clark2019SentenceMS}
Elizabeth Clark, Asli Celikyilmaz, and Noah~A. Smith. 2019.
\newblock Sentence mover's similarity: Automatic evaluation for multi-sentence
  texts.
\newblock In \emph{ACL}.

\bibitem[{Durmus et~al.(2020)Durmus, He, and Diab}]{durmus-etal-2020-feqa}
Esin Durmus, He~He, and Mona Diab. 2020.
\newblock \href {https://doi.org/10.18653/v1/2020.acl-main.454} {{FEQA}: A
  question answering evaluation framework for faithfulness assessment in
  abstractive summarization}.
\newblock In \emph{Proceedings of the 58th Annual Meeting of the Association
  for Computational Linguistics}, pages 5055--5070, Online. Association for
  Computational Linguistics.

\bibitem[{Eyal et~al.(2019)Eyal, Baumel, and Elhadad}]{eyal-etal-2019-question}
Matan Eyal, Tal Baumel, and Michael Elhadad. 2019.
\newblock \href {https://doi.org/10.18653/v1/N19-1395} {Question answering as
  an automatic evaluation metric for news article summarization}.
\newblock In \emph{Proceedings of the 2019 Conference of the North {A}merican
  Chapter of the Association for Computational Linguistics: Human Language
  Technologies, Volume 1 (Long and Short Papers)}, pages 3938--3948,
  Minneapolis, Minnesota. Association for Computational Linguistics.

\bibitem[{Fabbri et~al.(2021)Fabbri, Kryscinski, McCann, Socher, and
  Radev}]{Fabbri2020SummEvalRS}
A.~R. Fabbri, Wojciech Kryscinski, Bryan McCann, R.~Socher, and Dragomir Radev.
  2021.
\newblock Summeval: Re-evaluating summarization evaluation.
\newblock \emph{Trans. Assoc. Comput. Linguistics}, 9:391--409.

\bibitem[{Falke et~al.(2019{\natexlab{a}})Falke, Ribeiro, Utama, Dagan, and
  Gurevych}]{Falke2019RankingGS}
Tobias Falke, Leonardo F.~R. Ribeiro, Prasetya~Ajie Utama, Ido Dagan, and Iryna
  Gurevych. 2019{\natexlab{a}}.
\newblock Ranking generated summaries by correctness: An interesting but
  challenging application for natural language inference.
\newblock In \emph{ACL}.

\bibitem[{Falke et~al.(2019{\natexlab{b}})Falke, Ribeiro, Utama, Dagan, and
  Gurevych}]{falke-etal-2019-ranking}
Tobias Falke, Leonardo F.~R. Ribeiro, Prasetya~Ajie Utama, Ido Dagan, and Iryna
  Gurevych. 2019{\natexlab{b}}.
\newblock \href {https://doi.org/10.18653/v1/P19-1213} {Ranking generated
  summaries by correctness: An interesting but challenging application for
  natural language inference}.
\newblock In \emph{Proceedings of the 57th Annual Meeting of the Association
  for Computational Linguistics}, pages 2214--2220, Florence, Italy.
  Association for Computational Linguistics.

\bibitem[{Gliwa et~al.(2019)Gliwa, Mochol, Biesek, and
  Wawer}]{Gliwa2019SAMSumCA}
Bogdan Gliwa, Iwona Mochol, Maciej Biesek, and Aleksander Wawer. 2019.
\newblock Samsum corpus: A human-annotated dialogue dataset for abstractive
  summarization.
\newblock \emph{ArXiv}, abs/1911.12237.

\bibitem[{Goodrich et~al.(2019)Goodrich, Rao, Saleh, and
  Liu}]{Goodrich2019AssessingTF}
B.~Goodrich, V.~Rao, Mohammad Saleh, and Peter~J. Liu. 2019.
\newblock Assessing the factual accuracy of generated text.
\newblock \emph{Proceedings of the 25th ACM SIGKDD International Conference on
  Knowledge Discovery \& Data Mining}.

\bibitem[{Graham(2015)}]{graham-2015-evaluating}
Yvette Graham. 2015.
\newblock \href {https://doi.org/10.18653/v1/D15-1013} {Re-evaluating automatic
  summarization with {BLEU} and 192 shades of {ROUGE}}.
\newblock In \emph{Proceedings of the 2015 Conference on Empirical Methods in
  Natural Language Processing}, pages 128--137, Lisbon, Portugal. Association
  for Computational Linguistics.

\bibitem[{Guo et~al.(2018)Guo, Pasunuru, and Bansal}]{guo-etal-2018-soft}
Han Guo, Ramakanth Pasunuru, and Mohit Bansal. 2018.
\newblock \href {https://doi.org/10.18653/v1/P18-1064} {Soft layer-specific
  multi-task summarization with entailment and question generation}.
\newblock In \emph{Proceedings of the 56th Annual Meeting of the Association
  for Computational Linguistics (Volume 1: Long Papers)}, pages 687--697,
  Melbourne, Australia. Association for Computational Linguistics.

\bibitem[{Hashimoto et~al.(2019)Hashimoto, Zhang, and
  Liang}]{Hashimoto2019UnifyingHA}
T.~Hashimoto, Hugh Zhang, and Percy Liang. 2019.
\newblock Unifying human and statistical evaluation for natural language
  generation.
\newblock \emph{ArXiv}, abs/1904.02792.

\bibitem[{Huang et~al.(2020)Huang, Wu, and Wang}]{huang-etal-2020-knowledge}
Luyang Huang, Lingfei Wu, and Lu~Wang. 2020.
\newblock \href {https://doi.org/10.18653/v1/2020.acl-main.457} {Knowledge
  graph-augmented abstractive summarization with semantic-driven cloze reward}.
\newblock In \emph{Proceedings of the 58th Annual Meeting of the Association
  for Computational Linguistics}, pages 5094--5107, Online. Association for
  Computational Linguistics.

\bibitem[{Kryscinski et~al.(2019)Kryscinski, McCann, Xiong, and
  Socher}]{Kryscinski2019EvaluatingTF}
Wojciech Kryscinski, B.~McCann, Caiming Xiong, and R.~Socher. 2019.
\newblock Evaluating the factual consistency of abstractive text summarization.
\newblock \emph{ArXiv}, abs/1910.12840.

\bibitem[{Landis and Koch(1977)}]{Landis1977TheMO}
J.~Landis and G.~Koch. 1977.
\newblock The measurement of observer agreement for categorical data.
\newblock \emph{Biometrics}, 33 1:159--74.

\bibitem[{Lin(2004)}]{lin-2004-rouge}
Chin-Yew Lin. 2004.
\newblock \href {https://www.aclweb.org/anthology/W04-1013} {{ROUGE}: A package
  for automatic evaluation of summaries}.
\newblock In \emph{Text Summarization Branches Out}, pages 74--81, Barcelona,
  Spain. Association for Computational Linguistics.

\bibitem[{Liu and Lapata(2019)}]{lapata2019}
Yang Liu and Mirella Lapata. 2019.
\newblock \href {http://arxiv.org/abs/1905.13164} {Hierarchical transformers
  for multi-document summarization}.
\newblock \emph{ACL}.

\bibitem[{Massarelli et~al.(2019)Massarelli, Petroni, Piktus, Ott,
  Rockt{\"a}schel, Plachouras, Silvestri, and Riedel}]{Massarelli2019HowDS}
Luca Massarelli, F.~Petroni, Aleksandra Piktus, Myle Ott, Tim Rockt{\"a}schel,
  Vassilis Plachouras, F.~Silvestri, and S.~Riedel. 2019.
\newblock How decoding strategies affect the verifiability of generated text.
\newblock \emph{ArXiv}, abs/1911.03587.

\bibitem[{Miller(1995)}]{10.1145/219717.219748}
George~A. Miller. 1995.
\newblock \href {https://doi.org/10.1145/219717.219748} {Wordnet: A lexical
  database for english}.
\newblock \emph{Commun. ACM}, 38(11):39–41.

\bibitem[{Nallapati et~al.(2016)Nallapati, Zhou, Santos, Çaglar G{\"u}lçehre,
  and Xiang}]{Nallapati2016AbstractiveTS}
Ramesh Nallapati, Bowen Zhou, C.~D. Santos, Çaglar G{\"u}lçehre, and
  B.~Xiang. 2016.
\newblock Abstractive text summarization using sequence-to-sequence rnns and
  beyond.
\newblock In \emph{CoNLL}.

\bibitem[{Narayan et~al.(2018)Narayan, Cohen, and Lapata}]{xsum-emnlp}
Shashi Narayan, Shay~B. Cohen, and Mirella Lapata. 2018.
\newblock Don't give me the details, just the summary! {T}opic-aware
  convolutional neural networks for extreme summarization.
\newblock In \emph{Proceedings of the 2018 Conference on Empirical Methods in
  Natural Language Processing}, Brussels, Belgium.

\bibitem[{Novikova et~al.(2017)Novikova, Dusek, Curry, and
  Rieser}]{Novikova2017WhyWN}
Jekaterina Novikova, Ondrej Dusek, A.~Curry, and Verena Rieser. 2017.
\newblock Why we need new evaluation metrics for nlg.
\newblock In \emph{EMNLP}.

\bibitem[{Raffel et~al.(2019)Raffel, Shazeer, Roberts, Lee, Narang, Matena,
  Zhou, Li, and Liu}]{Raffel2019ExploringTL}
Colin Raffel, Noam Shazeer, Adam Roberts, Katherine Lee, Sharan Narang, Michael
  Matena, Yanqi Zhou, W.~Li, and Peter~J. Liu. 2019.
\newblock Exploring the limits of transfer learning with a unified text-to-text
  transformer.
\newblock \emph{ArXiv}, abs/1910.10683.

\bibitem[{Rajpurkar et~al.(2016)Rajpurkar, Zhang, Lopyrev, and
  Liang}]{Rajpurkar2016SQuAD10}
Pranav Rajpurkar, Jian Zhang, Konstantin Lopyrev, and Percy Liang. 2016.
\newblock Squad: 100, 000+ questions for machine comprehension of text.
\newblock \emph{ArXiv}, abs/1606.05250.

\bibitem[{Rankel et~al.(2013)Rankel, Conroy, Dang, and
  Nenkova}]{rankel-etal-2013-decade}
Peter~A. Rankel, John~M. Conroy, Hoa~Trang Dang, and Ani Nenkova. 2013.
\newblock \href {https://www.aclweb.org/anthology/P13-2024} {A decade of
  automatic content evaluation of news summaries: Reassessing the state of the
  art}.
\newblock In \emph{Proceedings of the 51st Annual Meeting of the Association
  for Computational Linguistics (Volume 2: Short Papers)}, pages 131--136,
  Sofia, Bulgaria. Association for Computational Linguistics.

\bibitem[{Reiter and Belz(2009)}]{reiter-belz-2009-investigation}
Ehud Reiter and Anja Belz. 2009.
\newblock \href {https://doi.org/10.1162/coli.2009.35.4.35405} {An
  investigation into the validity of some metrics for automatically evaluating
  natural language generation systems}.
\newblock \emph{Computational Linguistics}, 35(4):529--558.

\bibitem[{Scialom et~al.(2019)Scialom, Lamprier, Piwowarski, and
  Staiano}]{Scialom2019AnswersUU}
Thomas Scialom, Sylvain Lamprier, Benjamin Piwowarski, and Jacopo Staiano.
  2019.
\newblock Answers unite! unsupervised metrics for reinforced summarization
  models.
\newblock In \emph{EMNLP/IJCNLP}.

\bibitem[{Sellam et~al.(2020)Sellam, Das, and Parikh}]{sellam-etal-2020-bleurt}
Thibault Sellam, Dipanjan Das, and Ankur Parikh. 2020.
\newblock \href {https://doi.org/10.18653/v1/2020.acl-main.704} {{BLEURT}:
  Learning robust metrics for text generation}.
\newblock In \emph{Proceedings of the 58th Annual Meeting of the Association
  for Computational Linguistics}, pages 7881--7892, Online. Association for
  Computational Linguistics.

\bibitem[{Shazeer and Stern(2018)}]{Shazeer2018AdafactorAL}
Noam Shazeer and Mitchell Stern. 2018.
\newblock Adafactor: Adaptive learning rates with sublinear memory cost.
\newblock In \emph{ICML}.

\bibitem[{Shwartz et~al.(2020)Shwartz, West, Bras, Bhagavatula, and
  Choi}]{Shwartz2020UnsupervisedCQ}
Vered Shwartz, Peter West, Ronan~Le Bras, Chandra Bhagavatula, and Yejin Choi.
  2020.
\newblock Unsupervised commonsense question answering with self-talk.
\newblock \emph{EMNLP}.

\bibitem[{Vasilyev et~al.(2020)Vasilyev, Dharnidharka, and
  Bohannon}]{Vasilyev2020FillIT}
Oleg~V. Vasilyev, Vedant Dharnidharka, and J.~Bohannon. 2020.
\newblock Fill in the blanc: Human-free quality estimation of document
  summaries.
\newblock \emph{ArXiv}, abs/2002.09836.

\bibitem[{Wang et~al.(2020)Wang, Cho, and Lewis}]{wang-etal-2020-asking}
Alex Wang, Kyunghyun Cho, and Mike Lewis. 2020.
\newblock \href {https://doi.org/10.18653/v1/2020.acl-main.450} {Asking and
  answering questions to evaluate the factual consistency of summaries}.
\newblock In \emph{Proceedings of the 58th Annual Meeting of the Association
  for Computational Linguistics}, pages 5008--5020, Online. Association for
  Computational Linguistics.

\bibitem[{Zhang et~al.(2020)Zhang, Kishore, Wu, Weinberger, and
  Artzi}]{Zhang2020BERTScoreET}
Tianyi Zhang, V.~Kishore, Felix Wu, K.~Weinberger, and Yoav Artzi. 2020.
\newblock Bertscore: Evaluating text generation with bert.
\newblock \emph{ArXiv}, abs/1904.09675.

\bibitem[{Zhao et~al.(2020)Zhao, Cohen, and Webber}]{Zhao2020ReducingQH}
Z.~Zhao, Shay~B. Cohen, and B.~Webber. 2020.
\newblock Reducing quantity hallucinations in abstractive summarization.
\newblock \emph{ArXiv}, abs/2009.13312.

\bibitem[{Zhu et~al.(2020)Zhu, Hinthorn, Xu, Zeng, Zeng, Huang, and
  Jiang}]{zhu2020boosting}
Chenguang Zhu, William Hinthorn, Ruochen Xu, Qingkai Zeng, Michael Zeng,
  Xuedong Huang, and Meng Jiang. 2020.
\newblock \href {http://arxiv.org/abs/2003.08612} {Boosting factual correctness
  of abstractive summarization}.
\newblock \emph{ArXiv}, abs/2003.08612.

\end{thebibliography}
